\documentclass[11pt,a4paper]{article}
\usepackage[hyperref]{emnlp2018}
\usepackage{times}
\usepackage{latexsym}

\usepackage{url}

\usepackage{eqnarray}
\usepackage{bm}
\usepackage{eqnarray}
\usepackage{amssymb}
\usepackage{amsmath}
\usepackage{threeparttable}
\usepackage{multirow}
\usepackage{booktabs}
\usepackage{footnote}
\usepackage{graphicx}  
\usepackage{lipsum}
\aclfinalcopy 

\newcommand*\myat{{\fontfamily{ptm}\selectfont\small @}}

\title{Greedy Search with Probabilistic N-gram Matching \\ for Neural Machine Translation}

\def\first{$^1$}
\def\second{$^2$}
\def\comma{$^,$}
\def\star{$^*$}
\author{Chenze Shao\first\comma\second ~~~~~~~~ Yang Feng\first\comma\second\star ~~~~~~~~~ Xilin Chen\first\comma\second
\\
 \first Key Laboratory of Intelligent Information Processing
\\ Institute of Computing Technology, Chinese Academy of Sciences (ICT/CAS)
\\
{ \second {University of Chinese Academy of Science}} \\
{ \tt \small \{\href{mailto:shaochenze18z@ict.ac.cn}{shaochenze18z},\href{mailto:fengyang@ict.ac.cn}{fengyang},\href{mailto:xlchen@ict.ac.cn}{xlchen}\}\myat ict.ac.cn} \\}	
\date{}
\begin{document}
\maketitle
\newcommand\blfootnote[1]{%
\begingroup 
\renewcommand\thefootnote{}\footnote{#1}%
\addtocounter{footnote}{-1}%
\endgroup
}
\begin{abstract}
Neural machine translation (NMT) models are usually trained with the word-level loss using the teacher forcing algorithm, which not only evaluates the translation improperly but also suffers from exposure bias. Sequence-level training under the reinforcement framework can mitigate the problems of the word-level loss, but its performance is unstable due to the high variance of the gradient estimation. On these grounds, we present a method with a differentiable sequence-level training objective based on probabilistic n-gram matching which can avoid the reinforcement framework. In addition, this method performs greedy search in the training which uses the predicted words as context just as at inference to alleviate the problem of exposure bias. Experiment results on the NIST Chinese-to-English translation tasks show that our method significantly outperforms the reinforcement-based algorithms and achieves an improvement of 1.5 BLEU points on average over a strong baseline system.
\blfootnote{*Corresponding Author}
\end{abstract}

\section{Introduction}

Neural machine translation (NMT) \cite{kalchbrenner2013recurrent,cho2014learning,sutskever2014sequence,bahdanau2014neural} has now achieved impressive performance \cite{wu2016google,gehring2017convolutional,vaswani2017attention,hassan2018achieving,chen2018best,lample2018phrase} and draws more attention. NMT models are built on the encoder-decoder framework where the encoder network encodes the source sentence to distributed representations and the decoder network reconstructs the target sentence form the representations word by word.

Currently, NMT models are usually trained with the word-level loss (i.e., cross-entropy) under the teacher forcing algorithm \cite{williams1989learning}, which forces the model to generate translation strictly matching the ground-truth at the word level. However, in practice it is impossible to generate translation totally the same as ground truth. Once different target words are generated, the word-level loss cannot evaluate the translation properly, usually under-estimating the translation. In addition, the teacher forcing algorithm suffers from the {\em exposure bias} \cite{ranzato2015sequence} as it uses different inputs at training and inference, that is ground-truth words for the training and previously predicted words for the inference. \citet{kim2016sequence} proposed a method of sequence-level knowledge distillation, which use teacher outputs to direct the training of student model, but the student model still have no access to its own predicted words. Scheduled sampling(SS) \cite{bengio2015scheduled,venkatraman2015improving} attempts to alleviate the exposure bias problem through mixing ground-truth words and previously predicted words as inputs during training. However, the sequence generated by SS may not be aligned with the target sequence, which is inconsistent with the word-level loss. 

In contrast, sequence-level objectives, such as BLEU \cite{papineni2002bleu}, GLEU \cite{wu2016google}, TER \cite{snover2006study}, and NIST \cite{doddington2002automatic}, evaluate translation at the sentence or $n$-gram level and allow for greater flexibility, and thus can mitigate the above problems of the word-level loss. However, due to the non-differentiable of sequence-level objectives, previous works on sequence-level training \cite{ranzato2015sequence,shen2016minimum,bahdanau2016actor,wu2016google,he2016dual,wu2017adversarial,yang2017improving} mainly rely on reinforcement learning algorithms \cite{williams1992simple,sutton2000policy} to find an unbiased gradient estimator for the gradient update. Sparse rewards in this situation often cause the high variance of gradient estimation, which consequently leads to unstable training and limited improvements.

\citet{lamb2016professor,gu2017trainable,ma2018bag} respectively use the discriminator, critic and bag-of-words target as sequence-level training objectives, all of which are directly connected to the generation model and hence enable direct gradient update. However, these methods do not allow for direct optimization with respect to evaluation metrics.

In this paper, we propose a method to combine the strengths of the word-level and sequence-level training, that is the direct gradient update without gradient estimation from word-level training and the greater flexibility from sequence-level training. Our method introduces probabilistic $n$-gram matching which makes sequence-level objectives (e.g., BLEU, GLEU) differentiable. During training, it abandons teacher forcing and performs greedy search instead to take into consideration the predicted words. Experiment results show that our method significantly outperforms word-level training with the cross-entropy loss and sequence-level training under the reinforcement framework. The experiments also indicate that greedy search strategy indeed has superiority over teacher forcing.
\section{Background}
NMT is based on an end-to-end framework which directly models the translation probability from the source sentence $\bm{x}$ to the target sentence $\bm{\hat y}$:
\begin{equation}
\label{eq:prob}
P(\bm{\hat y}|\bm{x}) = \prod_{j=1}^{T}p(\hat y_j|\bm{\hat y_{<j}},\bm{x},\theta),
\end{equation}
where $T$ is the target length and $\theta$ is the model parameters. Given the training set $D = \{\rm{\mathrm{X}}^\mathrm{M},Y^{M}\}$ with $M$ sentences pairs, the training objective is to maximize the log-likelihood of the training data as
\begin{equation}
\begin{aligned}
\label{eq:mle}
&\bm{\theta} = \arg\max_{\theta}\{\mathcal{L}(\theta)\}\\
\mathcal{L}(\theta) = \sum_{m=1}^{M}&\sum_{j=1}^{l^m}\log(p(\hat y_j^m|\bm{\hat y_{<j}}^m,\bm{x}^m,\theta)),
\end{aligned}
\end{equation}
where the superior $m$ indicates the m-th sentence in the dataset and $l^m$ is the length of m-th target sentence. 

In the above model, the probability of each target word $p(\hat y_j^m|\bm{\hat y_{<j}^m},\bm{x^m},\theta)$ is conditioned on the previous target words. The scenario is that in the training time, the teacher forcing algorithm is employed and the ground truth words from the target sentence are fed as context, while during inference, the ground truth words are not available and the previous predicted words are instead fed as context. This discrepancy is called {\em exposure bias}.

\section{Model}
\subsection{Sequence-Level Objectives}
Many automatic evaluation metrics of machine translation, such as BLEU, GLEU and NIST, are based on the n-gram matching. Assuming that $\bm{y}$ and $\bm{\hat y}$ are the output sentence and the ground truth sentence with length $T$ and $T'$ respectively, the count of an $n$-gram $\bm{g} = (g_1,\dots,g_n)$ in sentence $\bm{y}$ is calculated as
\begin{equation}
\label{eq:count}
\text{C}_{\bm{y}}(\bm{g}) =\sum_{t=0}^{T-n}\prod_{i=1}^{n}1{\{g_i = y_{t+i}\}},
\end{equation}
where $1\{\cdot\}$ is the indicator function. The matching count of the $n$-gram $\bm{g}$ between $\bm{\hat y}$ and $\bm{y}$ is given by
\begin{equation}
\begin{aligned}
\label{eq:clip}
\text{C}^{\bm{\hat y}}_{\bm{y}}(\bm{g}) = \min{(\text{C}_{\bm{y}}(\bm{g})}, \text{C}_{\bm{\hat y}}(\bm{g})).
\end{aligned}
\end{equation}
Then the precision $p_n$ and the recall $r_n$ of the predicted $n$-grams are calculated as follows
\begin{eqnarray}
\label{eq:precision}
p_n = \frac{\sum_{\bm{g} \in \bm{y}}\text{C}^{\bm{\hat y}}_{\bm{y}}(\bm{g})}{\sum_{\bm{g}\in \bm{y}}\text{C}_{\bm{y}}(\bm{g})},\\
r_n = \frac{\sum_{\bm{g} \in \bm{y}}\text{C}^{\hat y}_{\bm{y}}(\bm{g})}{\sum_{\bm{g} \in \bm{\hat y}}\text{C}_{\bm{\hat y}}(\bm{g})}.
\end{eqnarray}

BLEU, the most widely used metric for machine translation evaluation, is defined based on the n-gram precision as follows
\begin{equation}
\label{eq:bleu}
\text{BLEU} = \text{BP} \cdot \exp(\sum_{n=1}^{N} w_n \log p_n),
\end{equation}
where BP stands for the brevity penalty and $w_n$ is the weight for the $n$-gram. In contrast, GLEU is the minimum of recall and precision of $1$-$4$ grams where $1$-$4$ grams are counted together:
\begin{equation}
\label{eq:gleu}
\text{GLEU} = \min(p_{1\text{-}4},r_{1\text{-}4}).
\end{equation}

\subsection{probabilistic Sequence-Level Objectives}

\begin{figure*}
  \begin{center}
    \includegraphics[scale=0.4]{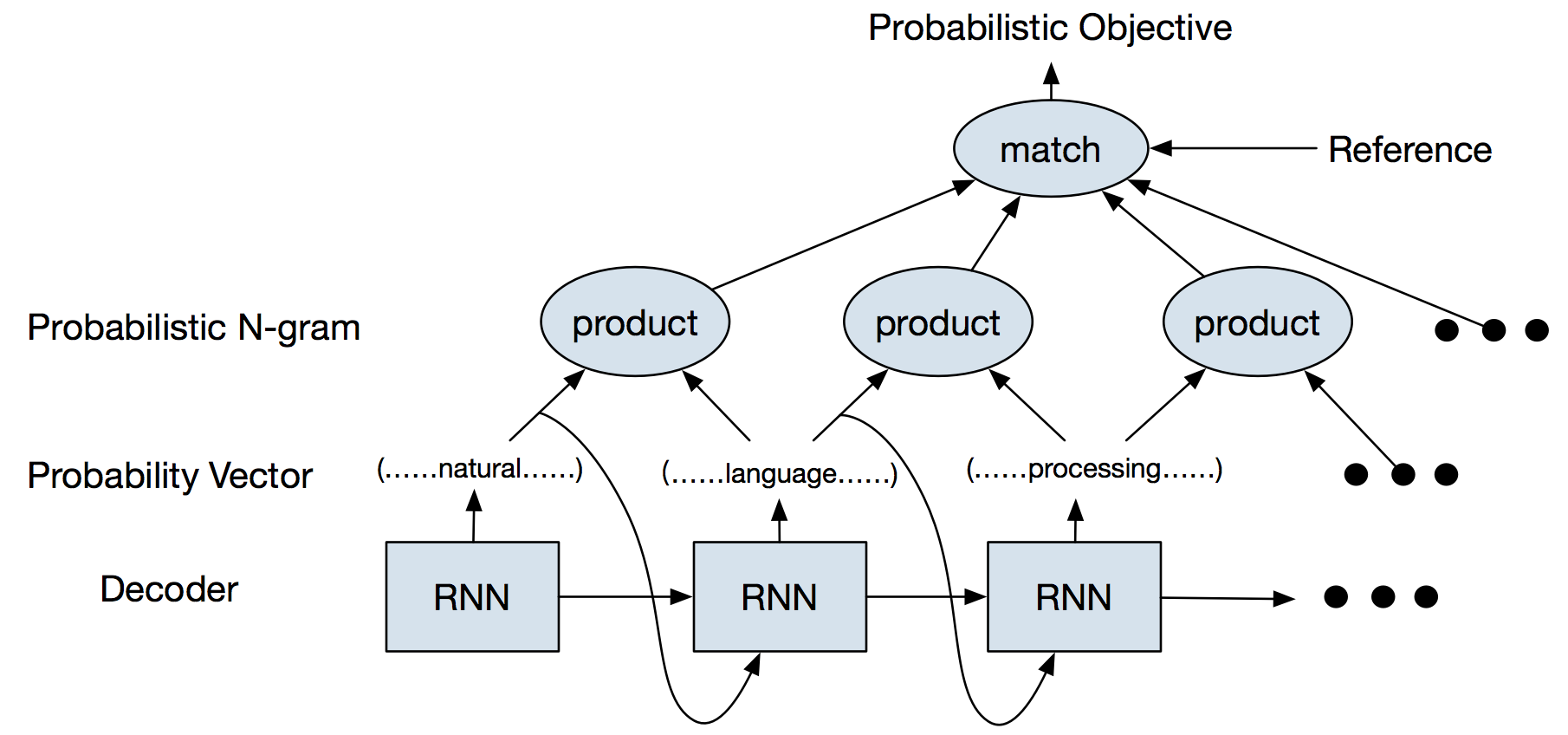}
    \caption{The overview of our model with greedy search. At each decoding step, the predicted word which has the highest probability in the probability vector is selected as context and fed into the RNN, and meanwhile this word and its probability are also used to calculate the probabilistic $n$-gram count.}
    \label{fig:graph}
  \end{center}
  \vspace{-0.5em}
\end{figure*}

In the output sentence $\bm{y}$, the prediction probability varies among words. Some words are translated by the model with high confidence while some words are translated with high uncertainty. However, when calculating the count of $n$-grams in Eq.(\ref{eq:count}), all the words in the output sentence are treated equally, regardless of their respective prediction probabilities. 

To give a more precise description of $n$-gram counts which considers the variety of prediction probabilities, we use the prediction probability $p(y_j|\bm{y_{<j}},\bm{x},\theta)$ as the count of word $y_j$, and correspondingly the count of an n-gram is the product of these probabilistic counts of all the words in the $n$-gram, not {\em one} anymore. Then the probabilistic count of $\bm{g} = (g_1,\dots,g_n)$ is calculated by summing over the output sentence $\bm{y}$ as
\begin{equation}
\begin{aligned}
\label{eq:wcount}
&\widetilde{\text{C}}_{\bm{y}}(\bm{g}) = \\
&\sum_{t=0}^{T-n}\prod_{i=1}^{n}1{\{g_i = y_{t+i}\}}\cdot p(y_{t+i}|\bm{y_{<t+i}},\bm{x},\theta).
\end{aligned}
\end{equation}

Now the probabilistic sequence-level objective can be got by replacing $\text{C}_{\bm{y}}(\bm{g})$ with $\widetilde{\text{C}}_{\bm{y}}(\bm{g})$ (the tilde over the head indicates the probabilistic version) and keeping the rest unchanged. Here, we take BLEU as an example and show how the probabilistic BLEU (denoted as P-BLEU) is defined. From this purpose, the matching count of n-gram $\bm{g}$ in Eq.(\ref{eq:clip}) is modified as follows
\begin{equation}
\begin{aligned}
\label{eq:wclip}
\widetilde{\text{C}}^{\bm{\hat y}}_{\bm{y}}(\bm{g}) = \min(\widetilde{\text{C}}_{\bm{y}}(\bm{g}), \text{C}_{\bm{\hat y}}(\bm{g})).
\end{aligned}
\end{equation}
and the predict precision of $n$-grams changes into
\begin{equation}
\label{eq:wprecision}
\tilde p_n = \frac{\sum_{\bm{g} \in \bm{y}}\widetilde{\text{C}}^{\hat y}_{\bm{y}}(\bm{g})}{\sum_{\bm{g} \in \bm{y}}\widetilde{\text{C}}_{\bm{y}}(\bm{g})}.
\end{equation}
Finally, the probabilistic BLEU (P-BLEU) is defined as
\begin{equation}
\label{eq:wbleu}
\text{P-BLEU} = \text{BP} \cdot \exp(\sum_{n=1}^{N} w_n \log \tilde p_n),
\end{equation}

Probabilistic GLEU (P-GLEU) can be defined in a similar way. Specifically, we denote the probabilistic precision of n-grams as P-Pn. The probabilistic precision is more reasonable than recall since the denominator in Eq.(\ref{eq:wprecision}) plays a normalization role, so we modify the definition in Eq.(\ref{eq:gleu}) and define P-GLEU as simply the probabilistic precision of 1-4 grams.

The general probabilistic loss function is:
\begin{equation}
\label{eq:obj}
\mathcal{L}(\theta) = -\sum_{m=1}^{M} \mathcal{P}(\bm{y}^m,\hat{\bm{y}}^m),
\end{equation}
where $\mathcal{P}$ represents the probabilistic sequence-level objectives, and $\bm{y}^m$ and $\hat{\bm{y}}^m$ are the predicted translation and the ground truth for the $m$-th sentence respectively. The calculation of the probabilistic objective is illustrated in Figure \ref{fig:graph}. This probabilistic loss can work with decoding strategies such as greedy search and teacher forcing. In this paper we employ greedy search rather than teacher forcing so as to use the previously predicted words as context and alleviate the exposure bias problem.
\begin{table*}[!htbp]
\centering
\begin{tabular}{c|c|c|c|c|c|c}
\toprule
\textbf{System}&\textbf{Dev(MT02)}&\textbf{MT03}&\textbf{MT04}&\textbf{MT05}&\textbf{MT06}&\textbf{AVG}\\
\midrule
\hline
BaseNMT&36.72&33.95&37.44&33.96&33.09&34.61\\
\hline
MRT&37.17&34.89&37.90&34.62&33.78&35.30\\
\hline
RF&37.13&34.66&37.69&34.55&33.74&35.16\\
\hline
P-BLEU&37.26&34.54&38.05&34.30&34.11&35.25\\
\hline
P-GLEU&37.44&34.67&38.11&34.24&34.58&35.40\\
\hline
P-P2&\textbf{38.03}&\textbf{35.45}&\textbf{39.30}&\textbf{35.10}&\textbf{34.59}&\textbf{36.11}\\
\bottomrule
\end{tabular}
\caption{Results on NIST Chinese-to-English Translation Task. \textbf{AVG} = average BLEU scores for test sets. The bold number indicates the highest score in the column.}
\label{tab1}
\end{table*}
\section{Experiment}
\subsection{Settings}
We carry out experiments on Chinese-to-English translation.\footnote{Experiment code: https://github.com/ictnlp/GS4NMT} The training data consists of 1.25M pairs of sentences extracted from LDC corpora\footnote{The corpora include LDC2002E18, LDC2003E07, LDC2003E14, Hansards portion of LDC2004T07, LDC2004T08 and LDC2005T06.}. Sentence pairs with either side longer than 50 were dropped. We use NIST 2002 (MT 02) as the validation set and NIST 2003-2006 (MT 03-08) as the test sets. We use the case insensitive 4-gram NIST BLEU score \cite{papineni2002bleu} for the translation task.

We apply our method to an attention-based NMT system \cite{bahdanau2014neural} implemented by Pytorch. Both source and target vocabularies are limited to 30K. All word embedding sizes are set to 512, and the sizes of hidden units in both encoder and decoder RNNs are also set to 512. All parameters are initialized by uniform distribution over $\left[-0.1,0.1\right]$. The mini-batch stochastic gradient descent (SGD) algorithm is employed to train the model with batch size of 40. In addition, the learning rate is adjusted by adadelta optimizer \cite{zeiler2012adadelta} with $\rho=0.95$ and $\epsilon=1e\textnormal{-}6$. Dropout is applied on the output layer with dropout rate of 0.5. The beam size is set to 10.

\subsection{Performance}

\textbf{Systems} We first pretrain the baseline model by maximum likelihood estimation (MLE) and then refine the model using probabilistic sequence-level objectives, including P-BLEU, P-GLEU and P-P2 (probabilistic 2-gram precision). In addition, we reproduce previous works which train the NMT model through minimum risk training (MRT) \cite{shen2016minimum} and REINFORCE algorithm (RF) \cite{ranzato2015sequence}. When reproducing their works, we set BLEU, GLEU and 2-gram precision as training objectives respectively and find out that GLEU yields the best performance. In the following, we only report the results with training objective GLEU.\\
\textbf{Performance} Table \ref{tab1} shows the translation performance on test sets measured in BLEU score. Simply training NMT model by the probabilistic 2-gram precision achieves an improvement of 1.5 BLEU points, which significantly outperforms the reinforcement-based algorithms. We also test the precision of other n-grams and their combinations, but do not notice significant improvements over P-P2. Notice that our method only changes the loss function, without any modification on model structure and training data. 

\subsection{Why Pretraining}
We use the probabilistic loss to finetune the baseline model rather than training from scratch. This is in line with our motivation: to alleviate the exposure bias and make the model exposed to its own output during training. 
In the very beginning of the training, the model's translation capability is nearly zero and the generated sentences are often meaningless and do not contain useful information for the training, so it is unreasonable to directly apply the greedy search strategy. Therefore, we first apply the teacher forcing algorithm to pretrain the model, and then we let the model generate the sentences itself and learn from its own outputs.

Another reason favoring pretraining is that pretraining can lower the training cost. The training cost of the introduced probabilistic loss is about three times higher than the cost of cross entropy. Without pretraining, the training time will be much higher than usual. Otherwise, the training cost is acceptable if the probabilistic loss is only for finetuning. 
\subsection{Effect of Decoding Strategy}

The probabilistic loss, defined in Eq.(\ref{eq:obj}), is computed from the model output $\bm{y}$ and reference $\hat{\bm{y}}$. In this section, we apply two different decoding strategies to generate $\bm{y}$: 1. teacher forcing, which uses the ground truth as decoder input. 2. greedy search, which feeds the word with maximum probability. By conducting this experiment, we attempt to figure out where the improvements come from: the modification of loss or the mitigation of exposure bias?

\begin{figure}
  \begin{center}
    \includegraphics[scale=0.5]{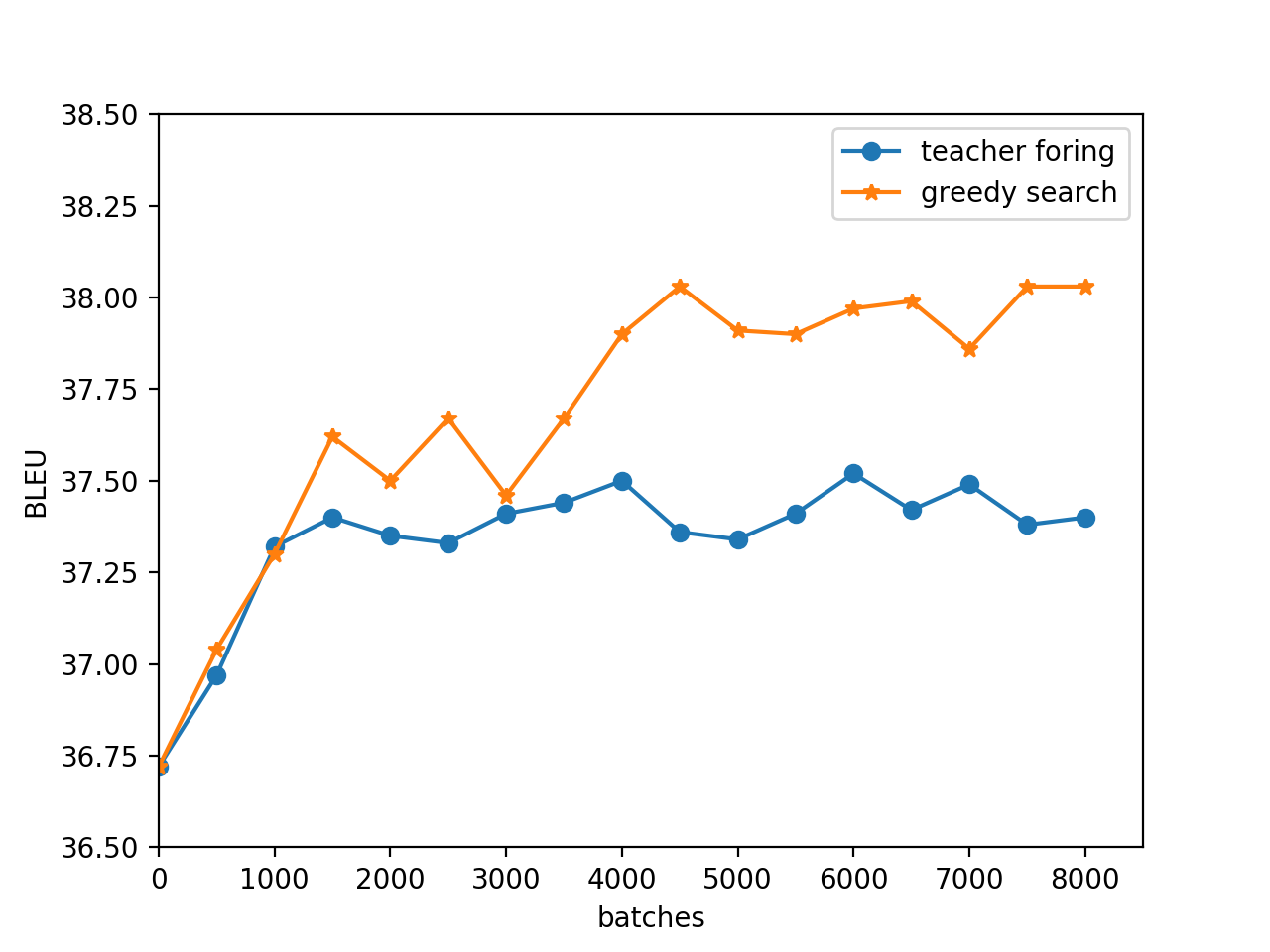}
    \caption{learning curves of different decoding strategies with training objective P-P2. }
    \label{fig:strategy}
  \end{center}
  \vspace{-0.5em}
\end{figure}

Figure \ref{fig:strategy} shows the learning curves of the two decoding strategies with training objective P-P2. Teacher forcing raises about 0.5 BLEU improvements and greedy search outperform the teacher forcing algorithm by nearly 1 BLEU point. We conclude that the probabilistic loss has its own advantage even when trained by the teacher forcing algorithm, and greedy search is effective in alleviating the exposure bias.

Notice that the greedy search strategy highly relys on the probabilistic loss and can not be conducted independently. Greedy search together with the word-level loss is very similar with the scheduled sampling(SS). However, SS is inconsistent with the word-level loss since the word-level loss requires strict alignment between hypothesis and reference, which can only be accomplished by the teacher forcing algorithm.

\subsection{Correlation with Evaluation Metrics}
In this section, we explore how the probabilistic objective correlates with the real evaluation metric. We randomly sample 100 pairs of sentences from the training set and compute their P-GLEU and GLEU scores (\citet{wu2016google} indicates that GLEU have better performance in the sentence-level evaluation than BLEU). 

\begin{figure}
  \begin{center}
    \includegraphics[scale=0.5]{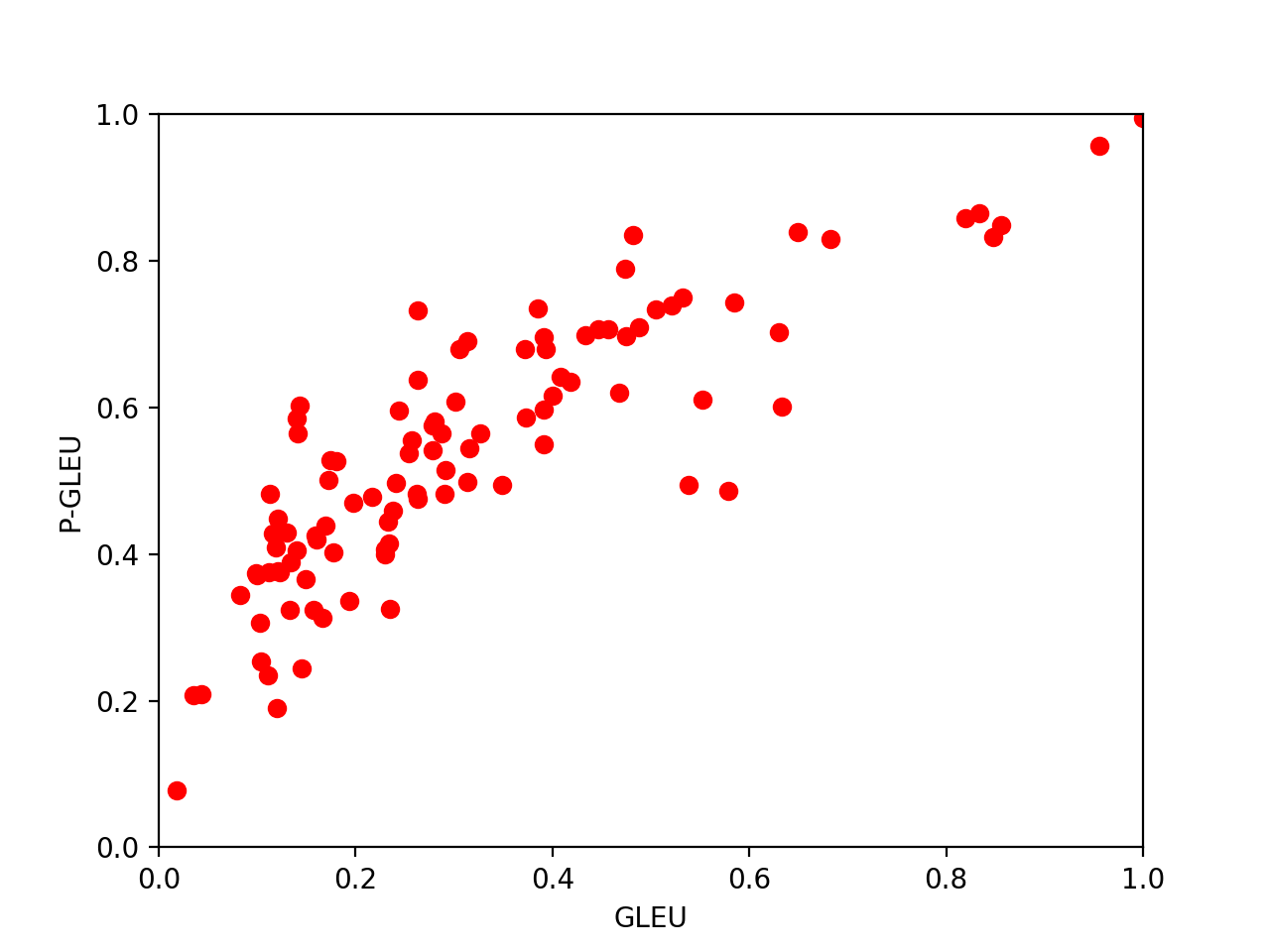}
    \caption{P-GLEU and GLEU scores on 100 pairs of sentences.}
    \label{fig:corr}
  \end{center}
  \vspace{-0.5em}
\end{figure}

Directly computing the correlation between GLEU and P-GLEU gives the correlation coefficient 0.86, which indicates strong correlation. In addition, we draw the scatter diagram of the 100 pairs of sentences in Figure \ref{fig:corr} with GLEU as x-axis and P-GLEU as y-axix. Figure \ref{fig:corr} shows that P-GLEU correlates well with GLEU, suggesting that it is reasonable to directly train the NMT model with P-GLEU. 
\section{Conclusion}
Word-level loss cannot evaluate the translation properly and suffers from the exposure bias, and sequence-level objectives are usually indifferentiable and require gradient estimation. We propose probabilistic sequence-level objectives based on n-gram matching, which relieve the dependence on gradient estimation and can directly train the NMT model. Experiment results show that our method significantly outperforms previous sequence-level training works and successfully alleviates the exposure bias through performing greedy search.
\section{Acknowledgments}
We thank the anonymous reviewers for their insightful comments. This work was supported by the National Natural Science Foundation of China (NSFC) under the project NO.61472428 and the project NO. 61662077.
\bibliography{emnlp2018}
\bibliographystyle{acl_natbib_nourl}
\end{document}